*Article*
# Synthetic Aperture Anomaly Imaging

Rakesh John Amala Arokia Nathan and Oliver Bimber*

Johannes Kepler University Linz, Austria
* Correspondence: oliver.bimber@jku.at; Tel.: +43-732-2468-6631

**Abstract:** Previous research has shown that in the presence of foliage occlusion, anomaly detection performs significantly better in integral images resulting from synthetic aperture imaging compared to applying it to conventional aerial images. In this article, we hypothesize and demonstrate that integrating detected anomalies is even more effective than detecting anomalies in integrals. This results in enhanced occlusion removal, outlier suppression, and higher chances of visually as well as computationally detecting targets that are otherwise occluded. Our hypothesis was validated through both: simulations and field experiments. We also present a real-time application that makes our findings practically available for blue-light organizations and others using commercial drone platforms. It is designed to address use-cases that suffer from strong occlusion caused by vegetation, such as search and rescue, wildlife observation, early wildfire detection, and surveillance.

## 1. Introduction

Several time-critical aerial imaging applications, such as search and rescue (SAR), early wildfire detection, wildlife observation, border control, and surveillance are affected by occlusion caused by vegetation, particularly forests. With Airborne Optical Sectioning (AOS) [1-16], we have introduced a synthetic aperture imaging technique that removes occlusion in real-time (cf. Fig. 1). AOS utilizes conventional camera drones to capture a sequence of single images with telemetry data while flying along a path which defines an extremely wide synthetic aperture (SA). These images are then computationally registered and integrated (i.e., averaged) based on defined visualization parameters, such as a virtual focal plane. Aligning this focal plane with the forest floor, for instance, results in a shallow depth-of-field integral image of the ground surface. It approximates the image that a physically impossible optical lens of the SA's size would capture. In this integral image, the optical signals of out-ot-focus occluders are suppressed, while focused targets are emphasized.

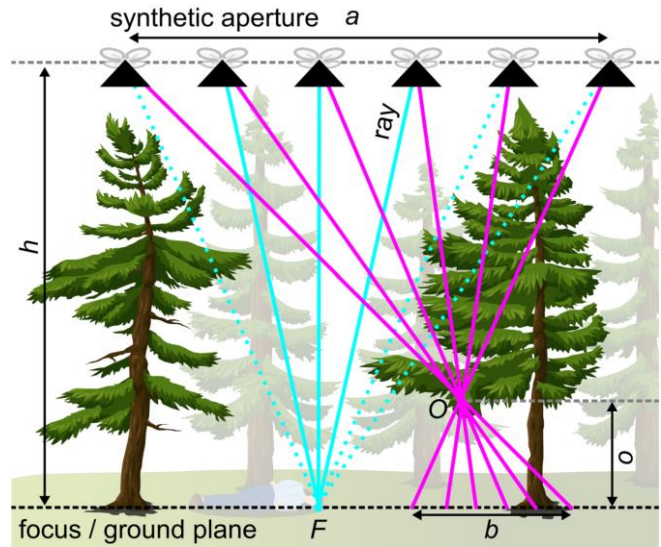

**Figure 1.** AOS principle: Registering and integrating multiple images captured along a synthetic aperture *a* while computationally focusing on focal plane *F* at distance *h* will defocus occluders *O* at distance *o* from *F* (with a point-spread of *b*) while focusing targets on *F*.

The principle synthetic aperture sensing applied by AOS in the optical field is also commonly used in other sensing domains where sensor size correlates to signal quality. The physical size limitations of sensors are overcome by computationally combining multiple measurements of small sensors to improve signal quality. This principle has found its application in various fields, including (but not limited) to radar [17-43], radio telescopes [44, 45], interferometric microscopy [46], sonar [47-50], ultrasound [51, 52], LiDAR [53, 54], and imaging [55-62].

One major advantage of AOS, in addition to its real-time capability, is its wavelength independence. The same technique can be used in the visible range, near-infrared range, and far-infrared range, which opens up many application fields. We have demonstrated that image processing techniques, such as deep-learning based classification [8-10] and anomaly detection [12,15], benefit significantly from integral images when compared to conventional single-image inputs. As a result, we have explored the application of AOS in search and rescue with autonomous drones [8-10], bird census in ornithology [5] and through-foliage tracking for surveillance and wildlife observation [12,14,16].

One advantage of model-based anomaly detection over machine-learning based classification is its robustness and invariance to training data. A common unsupervised anomaly detection method that can be applied to multispectral images is Reed–Xiaoli (RX) detection [63-64], which is often considered as benchmark for anomaly detection. It calculates global background statistics over the entire image and then compares individual pixels based on the Mahalanobis distance:

$$\alpha(r) = (r - \mu)^T K_{n \times n}^T (r - \mu),  \quad\quad\quad \text{(Eqn. 1)}$$

where $K_{n \times n}$ is the covariance matrix of the image with *n* input channels, the *n*-dimensional vector *r* is the pixel under test, and the *n*-dimensional vector $\mu$ is the image mean. The *t*% of all image pixels with the highest anomaly scores $\alpha$ are detected as abnormal by RX-detector, where *t* is referred to as RX threshold.

In [12], we have demonstrated that RX detection performs significantly better on integral images than on single images. The reason for this is that the background statistics of integral images is much more uniform than of single images, due to the defocused occluders. Our new hypothesis is that integrating anomalies (i.e., anomalies detected in single images before integration) even outperforms the detection of anomalies in integral images, as illustrated in Fig. 2. We refer to this principle as *Synthetic Aperture Anomaly Imaging* (SAAI).

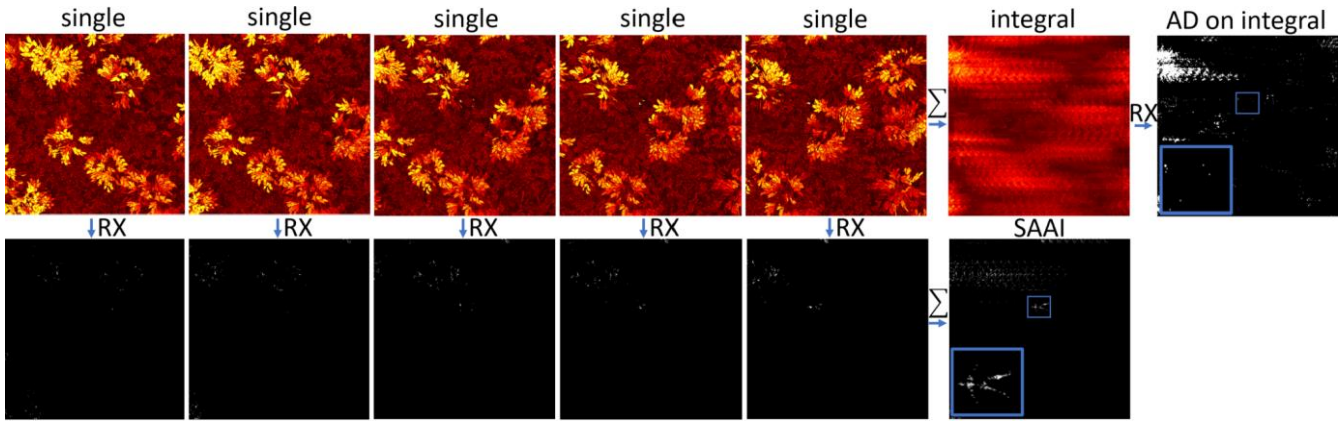

**Figure 2.** Top row: Integrating single images (color-mapped thermal images in this example) first and applying anomaly detection (AD) on the integral next. RX is used in this example. Bottom row: Applying anomaly detection to single images first, and then integrating the detected anomalies (SAAI). Results are simulated with a procedural forest model under sunny conditions (see Section 3.1).

We substantiate our hypothesis quantitatively through simulations that provide ground truth data. For real field experiments we have developed an application that is compatible with the latest DJI enterprise platforms, such as the Mavic 3T or the Matrice 30T. This application runs in real-time on DJI' Plus and Pro smart controllers and is freely available[1] to support blue-light organizations (BOS) and other organizations.

## 2. Results

The results presented in Fig. 3 are based on simulations conducted as described in Sect. 3.1, using procedural forests of different densities (300-500 trees/ha) with a hidden avatar lying on the ground. The far-infrared (thermal) channel is computed for cloudy and sunny environmental conditions, and is color-mapped (hot color bar, as shown in Fig. 2). We compare two cases: First, the thermal channel is integrated and the RX detector is applied to the integral (AD on integral), as done in [12]. Second, the RX detector is applied to the single images and the resulting anomalies are then integrated (SAAI).

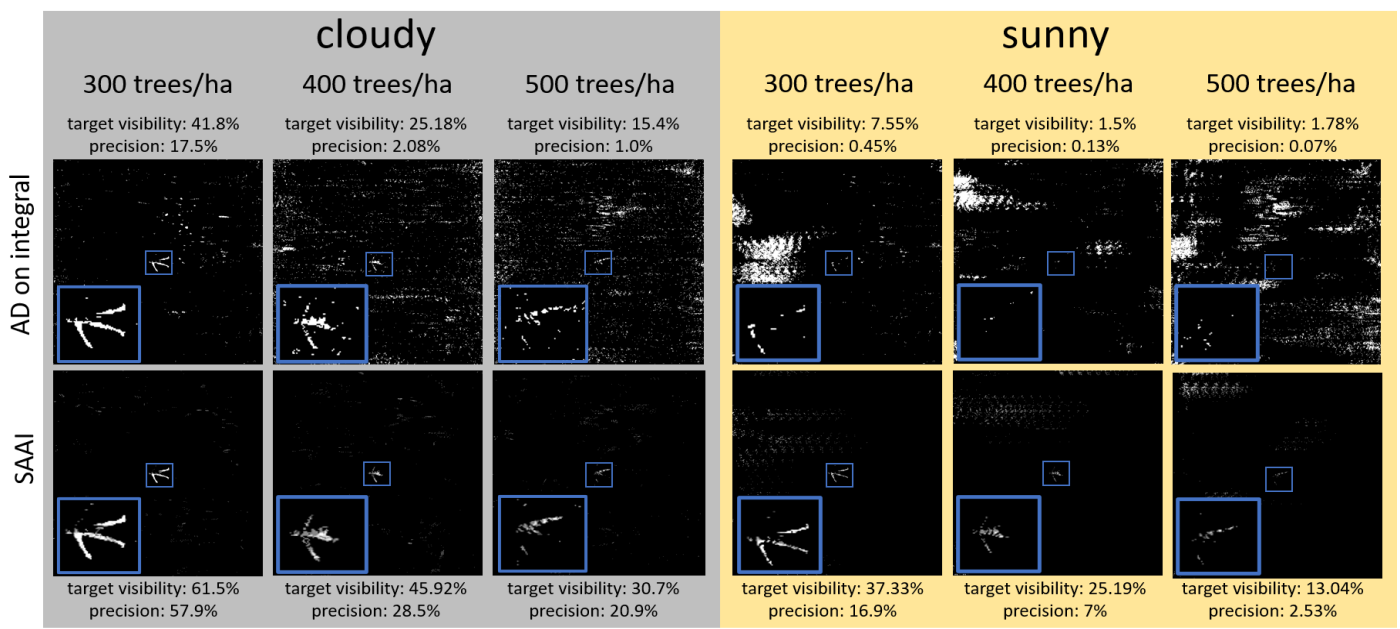

---

[1] https://github.com/JKU-ICG/AOS/

**Figure 3.** Simulated results for cloudy (left) and sunny (right) conditions and various forest densities (300, 400, 500 trees/ha): Anomaly detection applied to integral (top row) vs. integrated anomalies (bottom row). As shown in Fig. 2, color-mapped thermal images and RX detection were applied. We use a constant SA of $a$=10 m (integrating 10 images at 1 m sampling distance captured from an altitude of $h$=35 m AGL).

Since the ground truth projection of the unconcluded target can be computed in the simulation, its maximum visibility without occlusion is known. This is the aera of the target's projected footprint in the simulated images. One quality metric that can be considered is the remaining *target visibility* in case of occlusion. Thus, a target visibility of 61.5%, for instance, indicates that 61.5% of the complete target's footprint is still visible. However, target visibility considers only true positives (i.e., if a pixel that belongs to the target is visible or not). To consider false positives as well (i.e., pixels that are indicated but don't belong to the target), we determine *precision* as a second metric, which is the intensity integral of all true positives divided by the sum of true positives and false positives intensity integrals. High precision values indicate more true positives and fewer false positives.

As shown in Fig. 3, integrating anomalies (SAAI) always outperforms anomaly detections on integrals (AD on integral), both, in target visibility and precision. This is the case for all forest densities and also for cloudy and sunny environmental conditions. Target visibility and precision drop generally with higher densities due to more severe occlusion. In particular under sunny conditions, many and large false positive areas are detected. This is due to the higher thermal radiation of non-target objects, such as on tree-tops, that appear similar hot as the target. Under cloudy and cool conditions, the biggest thermal contribution comes mainly from the target itself.

One major difference between AD on integrals and SAAI is that the first case results in binary masks, as pixels are being indicated to be abnormal if they belong to the $t$% of pixels with highest anomaly scores $\alpha$ (Eqn. 1), while for the latter case the detections of the $t$% top-abnormal pixels of each single image are integrated. This leads to a non-binary value per pixels which corresponds to visibility (i.e., how often an abnormal point on the focal plane was captured free of occlusion).

Note that because the background models of single and integral images differ significantly [12], two different RX thresholds had to be applied to compare the two cases in Fig. 3 ($t$=99% for AD on integral, and $t$=90% for SAAI). They have been chosen such that the results of both cases approach each other as good as possible. For other thresholds SAAI outperforms AD on integral even more. Our simulations initially confirm our hypothesis that SAAI outperforms previous AD on integral approaches, and Sect. 4 discusses the reasons in more detail.

Figure 4 illustrates results of real field experiments under the more difficult (i.e., sunny) conditions for thermal imaging. For these experiments we implemented real-time SAAI on commercially available drones, as explained in Sect. 3.2. In contrast to simulated results, a ground truth does not exist here. Consequently, results can only be presented and compared visually.

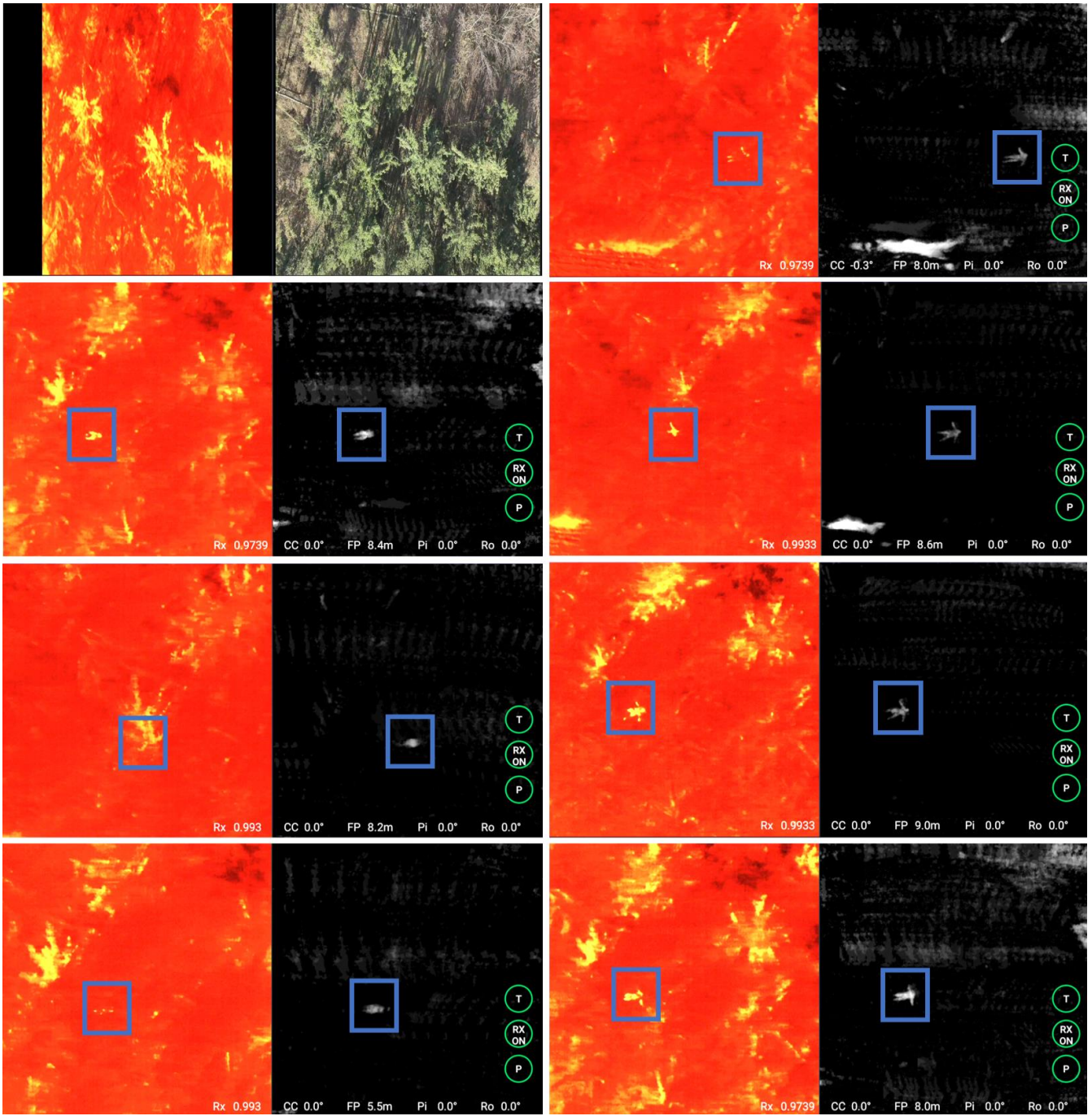

**Figure 4.** Real SAAI field-experiment results under sunny conditions: Screen-shots of our application running in real-time on the drone's smart controller. Color-mapped single thermal image and RGB image of our forest test site (top-left). SAAI detection results for sitting persons (left column) and for lying persons (right column). Left side of split-screen visualization shows the single image (color-mapped thermal) and ride side shows integral of anomalies (RX applied to single images before integration). See supplementary video for run-time details. We applied a maximal SA of $a$=15 m (integrating no more than 30 images at 0.5 m sampling distance, captured from an altitude of $h$=35 m AGL). Note, that depending on the local occlusion situations, the target became well visible after covering the max. SA to a shorter or larger extend. The RX threshold was individually optimized to achieve the best possible tradeoff between false and true positives ($t$=97.4% - 99.3%).

Compared to single thermal images where the target is partially or fully occluded and its fractional footprint is often indistinguishable from false heat singles of the surrounding, SAAI clearly reveals target shape and suppresses false heat signals well. Fig. 5 illustrates that applying anomaly detection to the integrals under the same conditions results in many wrong detections that form larger clusters, just like in our simulations. Identifying the target can be challenging for AD on integrals while SAAI leads to clear improvements.

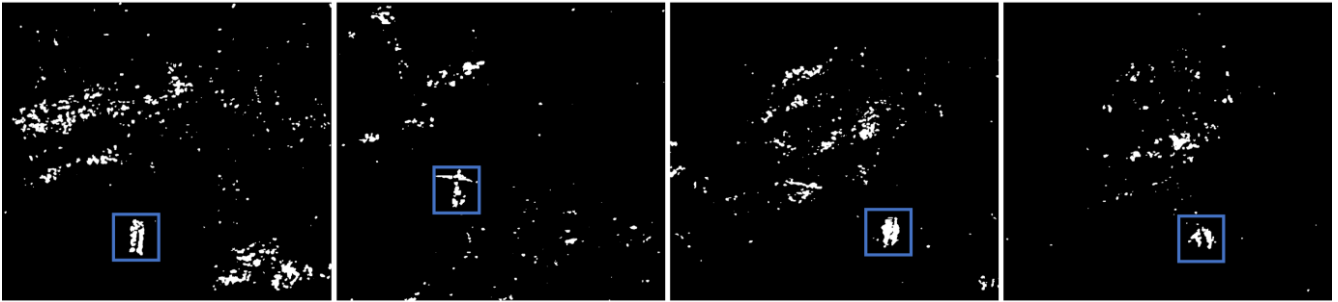

**Figure 5.** Real AD on integral field-experiment results under same conditions as in Fig. 5: Here, the RX detector is applied to the thermal integral images. Results show lying persons (left two images) and sitting persons (right two images). As in Fig.5, the RX threshold was individually optimized to achieve the best possible tradeoff between false and true positives ($t$=99.0% - 99.9%).

## 3. Materials and Methods

### 3.1. Simulation

Our simulations were realized with a procedural forest algorithm called *ProcTree*[2] and was implemented using WebGL. We computed 512x512 px aerial images (color-mapped thermal) for drone flights over a predefined area using defined sampling parameters (e.g., waypoints, altitudes, and camera field-of-view). Figure 1 shows examples of such simulated images. The virtual rendering camera (FOV = 50 deg in our case) applied perspective projection and was aligned with its look-at vector parallel to the ground surface normal (i.e., pointing downwards). Procedural tree parameters, such as tree height (20 m – 25 m), trunk length (4 m – 8 m), trunk radius (20 cm-50 cm), and leaf size (5 cm-20 cm) were used to generate a representative mixture of tree species. Finally, a seeded random generator was applied to generate a variety of trees at defined densities and degrees of similarity. Besides thermal effects of direct sunlight on tree crowns, other environmental properties, such as varying tree species, foliage and time of year, were assumed to be constant. The simulated forest densities were considered sparse with 300 trees/ha, medium with 400 trees/ha, and dense with 500 trees/ha. The simulated environment was a 1 ha procedural forest with one hidden avatar lying on the ground. Since the maximal visibility of the target (i.e., the maximal pixel coverage of the target's footprint in the simulated camera images under no occlusion) is known, the simulation allows quantitative comparisons in target visibility and precision, as presented in Fig. 3.

### 3.2. Real-Time Application on Drones

The real-time application used for our field-experiments was developed using DJI's Mobile SDK 5 which currently supports the new DJI enterprise series, such as Mavic 3T and Matric 30T. It runs on the Android 10 smart controllers (DJI RC Plus and Pro) and, as shown in Fig. 6, it supports three modes of operation: a *flight mode* that displays the live video stream of either the wide FOV, thermal, or zoom RGB camera, a *scan mode* and *parameter mode* in which single images for integration are recorded, the visualization parameters are interactively adjusted, and the live video stream (RGB or thermal) are displayed on the left side while the resulting integral (thermal, RGB, or RX = SAAI / AD on integral) is shown on the right side. The application supports networked Real-Time Kinematics (RTK), if available.

We forgo the use of soft-buttons or -sliders to be operated on the touch-screen. Instead, we map all functionality of the application to hard-buttons and -sticks. This allows more robust control and supports first-person-view (FPV) flying by attaching goggles to the HDMI port of the smart controller. The control assignments on the smart controller are chosen as follows (using DJI's naming convention): The C2 button switches forth and back between flight mode and scan mode. In parameter mode, it resets the focal plane parameters. The C1 button switches forth and back between scan mode and parameter mode. The record button switches forth and back between RGB and thermal imaging (and zoom-camera in flight mode only) in flight and scan modes. The shutter button turns on/off anomaly detection (RX). The C3 button opens parameter options (different settings for RGB and thermal imaging), such as RTK settings, sampling distance and integration window size, thermal color modes, etc. The left dial changes the gimbal tilt in flight and scan modes. The right dial changes the RX threshold or contrast of integral images (the pause button is used to toggle). In flight mode, it changes the zoom of the zoom camera. The pause button toggles between changing contrast of the integral image or RX threshold in parameter mode. The right stick controls the drone in flight and scan modes. In parameter mode, it changes the focal plane distance and compass correction settings (push/pull: focal plane up/down, left/right: compass correction counter-clockwise/clockwise). The left stick controls the drone in flight and scan modes. In parameter mode, it changes the focal plane orientation (push/pull: tilt forward/backward, left/right: tilt left/right).

---

[2] https://github.com/supereggbert/proctree.js

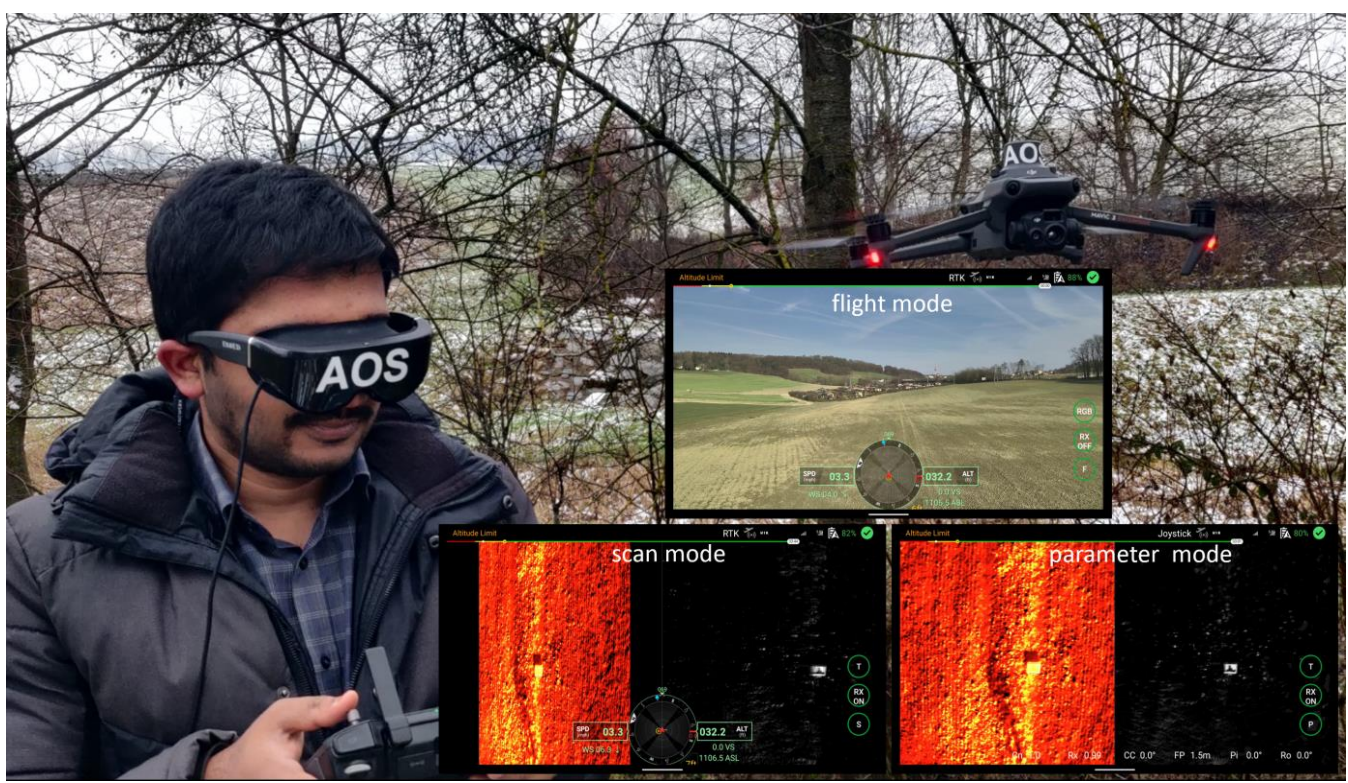

**Figure 6.** Common process of operation: (1) Take off and fly to target area in *flight mode*. (2) Scan in target area by flying a linear sideways path in *scan mode*. (3) Fine-tune visualization parameters (focal plane, compass correction, contrast, RX threshold) in *parameter mode*. Steps 2 and 3 can be repeated to cover larger areas. Operation supports interactive visualization directly on the smart control as well as first person view (FPV) using additional goggles.

Figure 7 depicts a schematic overview over the application's main software-system components. It consists of three parallelly running threads that share image and telemetry data through queues. This is essential for buffering slight temporal differences in runtime of these threads. The first thread (implemented in Java) receives raw (RGB or thermal) video and telemetry data (RTK corrected GPS, compass direction, gimbal angles) from the drone, decodes the video data, and selects frames (video-telemetry pairs) based on the defined sampling distance. Thus, if the selected sampling distance is 0.5 m, for example, only frames at flight distance $\geq$ 0.5 m are selected and pushed into the queue. Note, that while video streams are delivered at 30 Hz, GPS sampling is limited to a maximum of 10 Hz (10 Hz with RTK, 5 Hz with conventional GPS). Therefore, SAAI results are displayed at a speed of 10 Hz / 5 Hz.

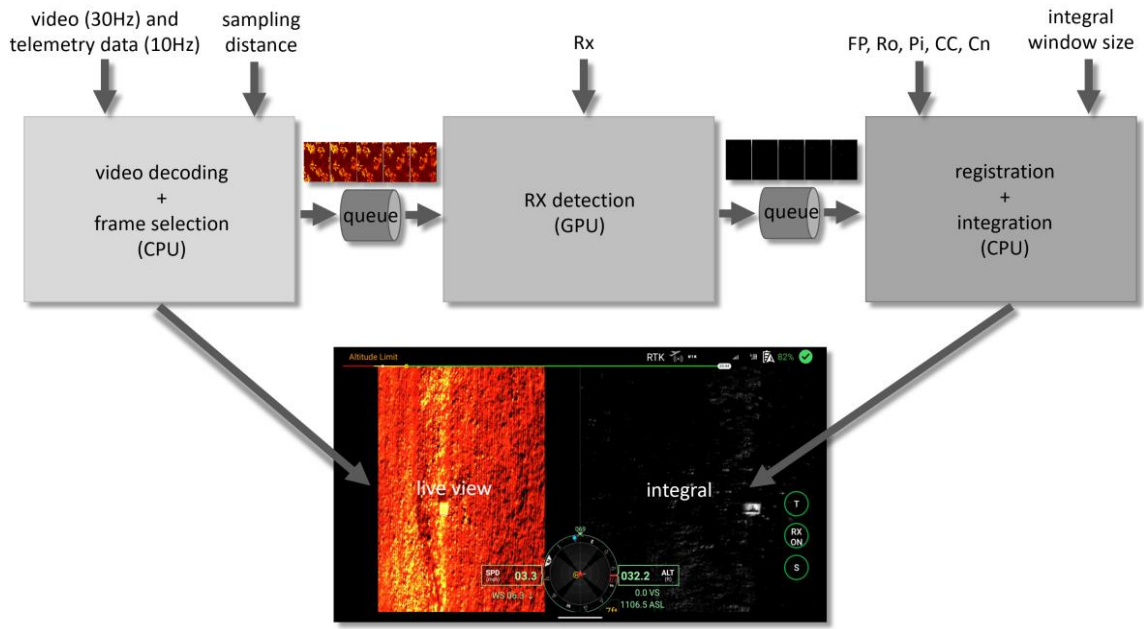

**Figure 7.** Software-system overview: The main components of our application are three parallel running threads that exchange video and telemetry data through two queues. They are distributed on CPU and GPU of the smart controller for optimal performance and parallel processing.

The next thread (implemented in PyTorch for C++) computes single image RX detection on the GPU of the smart controller. It receives the preselected frames from the first queue and computes the per-pixel anomalies base on the selected RX threshold $t$ (which is called *Rx* in the application's GUI). These frames are then forwarded through the second queue to the third thread. The last thread (implemented in C++) registers a window of images (i.e., the $n$ latest frames, where $n$ is the integral window size of the SA) based on their telemetry data and given visualization parameters: the focal plane distance $h$ (called *FP* in the application's GUI), pitch (*Pi*) and roll (*Ro*), the compass correction value (*CC*), and a contrast enhancement factor (*Cn*). The final integral is displayed on the right side of the split screen. Note, that *FP, Ro, Pi, CC, Cn,* and *Rx* can be interactively changed in parameter mode, as explained above. The *sampling distance* and *integral windows size* are defined in the application's settings. Note that *sampling distance* times *integral windows size* equals the SA size (*a* in Fig. 1).

Performance measures were timed on a DJI RC Pro Enterprise running on Android 10: The initial thread required 10 - 20 ms, the second thread 5 - 10 ms, and the last thread took approximately 40 ms. Overall, this led to approximately 45-75 ms required processing time. However, the maximum GPS sampling speed of 10Hz (RTK) defers the processing to 100 ms in practice.

## 4. Discussion and Conclusion

It was previously shown that, in the presence of foliage occlusion, anomaly detection performs significantly better in integral images resulting from synthetic aperture imaging than in conventional single aerial images [12]. The reason for this is the much more uniform background statistics of integral images compared to the one of single images. In this article, we demonstrate that integrating detected anomalies significantly outperforms detecting anomalies in integrals. This leads to enhanced occlusion removal and outlier suppression, and consequently, to higher chances for detecting otherwise occluded targets visually (i.e., by a human observer) as well as computationally (e.g., by an automatic classification method). This new finding can be explained as follows:

With respect to Fig. 1, the integral signal of a target point $F$ is the sum of all registered ray contributions of all overlapping areal images. This integral signal consists of a mixture of unconcluded (signal of target) and occluded (signal of forest background) ray contributions. Only if the unconcluded contributions dominate, the resulting integral pixel can robustly be detected as an anomaly. On the one hand, applying anomaly detection before integration zeros out occluding rays initially while assigning the highest possibly signal contribution of 1 to the target rays. Thus, integrating detected anomalies reduces background noise in the integrated target signal. In fact, the integrated target signal corresponds directly to visibility (i.e., how often an abnormal point on the focal plane was captured free of occlusion). On the other hand, the integral signals of occluders (e.g., *O* in Fig. 1) can also be high (e.g., tree crowns that are headed by sun light). They would be considered as anomalies if they differ too much from the background model, and would be binary masked just like targets. Consequently, anomaly detection applied to integrals can lead to severe false positives that are indistinguishable from true positives, as shown in Fig. 3 (top row). Certainly, this is also the case when anomaly detection is applied to single images. But integrating the anomaly masks suppresses the contribution of false positives as their rays are not registered on the focal plane, as shown in Fig. 3 (bottom row). Only false positives located directly on the focal plane (e.g., open ground patches that are heated by sunlight, as the large bright patches shown by the two top-right examples in Fig. 5) remain registered and can lead to classification confusion. For dense forests, however, we can expect that large open ground patches are rare, and that most of the sunlight that could cause false detections is reflected by the tree crowns.

We have already demonstrated that using more channels (e.g., RGB+thermal instead of RGB or thermal only) improves anomaly detection in general [15]. Since a simultaneous access of the raw RGB and thermal video streams is not supported by consumer DJI drones, we focused on the most relevant (thermal) channel in our application so far. Extending anomaly detection to more channels (including motion) is on our future work list. Also, the integration of classification approaches, such as the automatic person classifier that was trained for AOS data [10], will be considered in future. We expect that applying classification to SAAI results leads to better results than an application to integrals (as done in [10]), as false positive pixels are initially filtered out.

**Supplementary Materials:** Supporting information and the AOS application for DJI can be downloaded at: https://github.com/JKU-ICG/AOS/ (AOS for DJI). The supplementary video is available at: https://user-images.githubusercontent.com/83944465/217470172-74a2b272-2cd4-431c-9e21-b91938a340f2.mp4

**Data Availability Statement:** All experimental data presented in this article is available at: https://doi.org/10.5281/zenodo.7867080

**Author Contributions:** Conceptualization O.B.; methodology O.B. and R.N.; software R.N.; validation R.N. and O.B.; formal analysis R.N. and O.B.; investigation R.N.; resources R.N.; data curation R.N; writing—original draft preparation O.B.; writing—review and editing O.B. and R.N.; visualization O.B. and R.N.; supervision O.B.; project administration O.B.; funding acquisition O.B. All authors have read and agreed to the published version of the manuscript.

**Funding:** This research was funded by the Austrian Science Fund (FWF) and German Research Foundation (DFG) under grant numbers P 32185-NBL and I 6046-N, and by the State of Upper Austria and the Austrian Federal Ministry of Education, Science and Research via the LIT–Linz Institute of Technology under grant number LIT-2019-8-SEE-114.

**Acknowledgments:** We thank Francis Seits, Rudolf Ortner, and Indrajit Kurmi for contributing to the implementation of the application, and Mohammed Abbass for supporting the field experiments.

**Conflicts of Interest:** The authors declare no conflict of interest.